\UseRawInputEncoding
\documentclass[a4paper,twoside]{article}

\usepackage{epsfig}
\usepackage{subcaption}
\usepackage{calc}
\usepackage{amssymb}
\usepackage{amstext}
\usepackage{amsmath}
\usepackage{amsthm}
\usepackage{multicol}
\usepackage{pslatex}
\usepackage{apalike}
\usepackage[bottom]{footmisc}
\usepackage{SCITEPRESS}     

\begin{document}

\title{Fast Eye Detector Using Siamese Network for NIR Partial Face Images}

\author{\authorname{Yuka Ogino\sup{1}, Yuho Shoji\sup{1}, Takahiro Toizumi\sup{1}, Ryoma Oami\sup{1} and Masato Tsukada\sup{2}}
\affiliation{\sup{1}NEC Corporation, Kanagawa, Japan}
\affiliation{\sup{2}University of Tsukuba, Ibaraki, Japan}
\email{\{yogino, yuho-shoji, t-toizumi\_ct, r-oami\_az\}@nec.com, tsukada@iit.tsukuba.ac.jp}
}

\keywords{eye detection, object detection}

\abstract{
This paper proposes a fast eye detection method that is based on a Siamese network for near infrared (NIR) partial face images. NIR partial face images do not include the whole face of a subject since they are captured using iris recognition systems with the constraint of frame rate and resolution. The iris recognition systems such as the iris on the move (IOTM) system require fast and accurate eye detection as a pre-process. Our goal is to design eye detection with high speed, high discrimination performance between left and right eyes, and high positional accuracy of eye center. Our method adopts a Siamese network and coarse to fine position estimation with a fast lightweight CNN backbone. The network outputs features of images and the similarity map indicating coarse position of an eye. A regression on a portion of a feature with high similarity refines the coarse position of the eye to obtain the fine position with high accuracy. We demonstrate the effectiveness of the proposed method by comparing it with conventional methods, including SOTA, in terms of the positional accuracy, the discrimination performance, and the processing speed. Our method achieves superior performance in speed.
}

\onecolumn \maketitle \normalsize \setcounter{footnote}{0} \vfill

\vspace{-10pt}
\section{Introduction}
\begin{figure}[h]
\centering
\includegraphics[width=0.85\hsize]{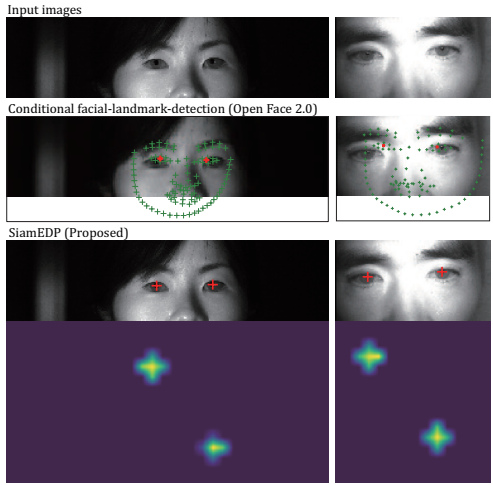}
\caption{Examples of NIR partial face images and detection results. First row shows input images. Left image was captured with iris on the move (IOTM) system, and right image is from public iris dataset CASIA-Iris-M1-S2 \cite{CASIAIrisM1}. Second row shows results of OpenFace2.0 \cite{baltrusaitis2018openface} (face detection + landmark detection). Third row shows results of proposed method. Last two rows show heat maps indicates right and left eye similarity of reference features generated from SiamEDP.}
\label{fig:introExamples}
\end{figure}

Iris and periocular recognition \cite{Daugman}, pupillometry \cite{wilhelm2003clinical,couret2016reliability}, and gaze tracking \cite{cheng2021appearance} are used for identifying individuals and human state estimation from near infrared (NIR) images. These methods require high resolution to capture the fine texture of irises from moving subject. In iris capturing systems, the field of view is limited to the both eye area only due to the constraint of resolution. CASIA-Iris-M1 \cite{CASIAIrisM1} and CASIA-Iris-Distance \cite{casiairisdistance} (examples in Fig. \ref{fig:introExamples}, and Fig. \ref{fig:DLibs}), public datasets for iris recognition, are good examples. These datasets were created using mobile devices and iris imaging systems at a distance (IAAD \cite{nguyen2017long}). These images are partial face images, which include both eyes but not an entire face. The iris-recognition process using these partial face images first detects and classifies a right eye and a left eye respectively, then extract iris regions by segmentation on single eye images, and finally extract features from them for authentication.

In particular, Iris On the Move (IOTM) \cite{matey2006iris} system, an iris authentication system for a walking individual, requires a high frame rate. High frame rate increases the chances of capturing focused irises from a walking individual passing through the narrow depth of field　reduced by the constraints of high resolution. Recent IOTM system \cite{zhang2020all} captures in 30 fps with 12M pixels. For real time processing of iris recognition in this system, the processing time for eye detection is required faster than 33 msec. The IOTM system requires a wide horizontal angle of view to expand width of walking pathway. The position of the face in the captured image is greatly shifted due to individual differences in gait and walking position, so the eyes do not always appear in the same position in images (examples in Fig. \ref{fig:introExamples}). A technique to detect each eyes with high speed and positional accuracy from NIR partial face images is expected.

In iris recognition, and certain gaze-tracking methods, precise eye landmark detectors \cite{choi2019accurate,ablavatski2020real,ahmed2021real,lee2020deep} have been proposed for estimating gaze direction and segmentation of the pupil and iris regions. These methods first detect the rectangle face region (bounding box) from the input image and crop the single eye regions from the bounding box using landmark detection. Landmark detection is a method of detecting a set of landmark points representing facial parts such as eyes, nose, and mouth. Finally, iris-landmark estimations and iris segmentations are executed for a single eye region. 

Current face-detection \cite{dalal2005histograms,king2015max,deng2020retinaface,bazarevsky2019blazeface,xiang2017joint} and facial-landmark-detection methods \cite{kazemi2014one,guo2019pfld,kartynnik2019real,zadeh2017convolutional} are fast and accurate. These detection methods improved performance of occlusions. These occlusions indicate that a face is shielded by objects such as a mask, or another person's face. In other words, the target image contains objects on the face, which is different from the partial face in which some parts of the face is out of the angle of view. Landmark detection methods does not assume the partial face image, since the input is a fixed resized face region obtained by face detection. Moreover, these detectors are trained using images captured under visible light and including whole face, so the pre-trained model cannot be used for NIR partial face images. Annotation of many landmark points such as facial bounding box, eyes, contours, eyebrows and nose, for a NIR partial-face dataset requires a great deal of effort. 

Direct eye-detection methods, such as object-detection methods, have also been proposed \cite{ren2015faster,yolov4}. Generic object detection methods \cite{liu2020deep} extract features from the input image and regress object classes and bounding boxes from the features to detect objects in different classes and different scales. These systems are not fast enough to meet the requirement for the real-time performance of the iris recognition system.

We proposes a fast eye detection method for NIR partial face images that is based on a Siamese network (SiamEDP) and directly detects right and left  eye centers. We focused on a fully-convolutional Siamese network \cite{SiamFC} (SiamFC) to accurately obtain the eye center with a lightweight model. The Siamese network extracts features from two kind of images. One is a NIR partial face image as a search image and the other is a single eye image prepared in advance as a reference image. SiamEDP outputs a coarse similarity heat map between the reference feature and the search feature. Classification by similarity using the reference features is expected to reduce training parameters, improve discriminative performance between left and right eyes, and stabilize training. We further extended the two-dimensional convolutional similarity to cosine-margin-based loss \cite{wang2018cosface} to improve the performance. SiamEDP regresses a search feature vector with a highest similarity, and obtain the local fine position of the eye center. Therefore, SiamEDP can detect coarse to fine eye-center positions with high speed and accuracy. We evaluated the accuracy of SiamEDP, and demonstrated the effectiveness of the Siamese network and cosine-margin-based loss. We also compared it with current facial-landmark-detection methods by using public iris-recognition datasets. The results indicate that SiamEDP is faster and more accurate than these two types of methods.

Our main contributions are as follows:
\vspace{-5pt}
\begin{itemize}
	\setlength{\itemsep}{3pt}
	\setlength{\parskip}{0pt}
   \item We propose an eye detector as a pre-process of iris segmentation for partially cropped NIR face image.
   \item We apply SiamFC\cite{SiamFC} for object tracking to eye detection with a light weight network.
   \item Using cosine-margin-based loss (CosFace\cite{wang2018cosface}) on training improves accuracy of detection.
   \item Coarse to fine approach improves positional accuracy.
   \item Our method reduces the cost of annotation less than facial-landmark detection. Only two landmarks of eye center are required for a single face.
\end{itemize}

\section{Related Work}
\subsection{Face and Landmark Based Eye Detection}
Eye-center or pupil-center detection methods have been proposed for gaze tracking \cite{cheng2021appearance}. These methods first extract the face region using face detection, then resize the region to a fixed size, extract the single eye regions using facial-landmark detection, and execute high-precision position estimation, or directly detect the single eye region from the face region. Therefore, the accuracy of these methods depends on the accuracy of the underlying face region and facial-landmark detection.

Face detection methods predict the facial bounding box. Early methods were mainly based on the classifiers using hand-crafted features extracted from an image \cite{violajones2001}. After the breakthrough of the CNN, CNN based models were proposed, such as Cascade-CNN, Faster-RCNN, and single-Shot Detection \cite{Li_2015_CVPR,Sun_2017,Zhang_2017,bazarevsky2019blazeface,deng2020retinaface}. To improve detection accuracy, several studies focused on the loss function or multi-task learning \cite{Ranjan2016HyperFace,deng2020retinaface}. Dent et al. \cite{deng2020retinaface} proposed RetinaFace which predicts facial bounding box by leveraging extra-supervised and self-supervised multi-task learning and showed significant improvement in accuracy. One of the challenges in face detection is occlusion, i.e, the lack of facial information due to obstacles or masks. Chen et al. \cite{Chen2018AOFD} proposed the Occlusion-aware Face Detector (AOFD) which detects faces with few exposed facial landmarks using adversarial training strategy.
The above face-detection methods use annotated facial-image datasets that include images captured under visible light. Several visible-light face datasets are publicly available. For example, WIDER FACE \cite{Yang2016WIDERFACE} includes more than 30,000 images and about 4 million labeled faces. There are several other datasets containing hundreds to tens of thousands of labeled faces. The majority of images are wide-angle shots of the face \cite{fddbTech,YAN2014790,Yang2015,Nada2018,Cao2018VGGFace2}. 

Several of eye-detector and eye-center estimation methods detect eyes from resized facial bounding boxes. Some methods \cite{ahmed2021real,leo2013unsupervised} use statistical facial-landmark information for cropping out single eyes before an eye segmentation process in real time. The other method \cite{putro2020fast} uses a face region resized to $128 \times 128$ pixels before a bounding-box estimation of eyes.

Facial-landmark-detection methods detect key points that represent facial landmarks from facial bounding boxes. Early landmark-detection methods were mainly based on fitting a deformable face mesh by using statistical methods \cite{WANG201850}. V. Kazemi et al, proposed ensemble of regression trees which is based on gradient boosting initialized with the mean shape of landmarks \cite{kazemi2014one}. They achieved high speed and high accuracy in detecting 68 points from frontal-face images with less occlusion. CNN based landmark detectors are also proposed, showing significant improvement in in-the-wild facial-landmark detection \cite{sun2013deep,zhou2013extensive,Chandran2020CVPR,Zhang2016joint,Feng2018CVPR}. 
The models of these methods are typically evaluated with 68 points using annotated visible-light-image datasets. Several datasets \cite{Sagonas2013_300Face,AFLW2011,Burgos2013COFW,Wu2018WFLW} are available. Each dataset includes several thousand of annotated images. 
For example, the 300W \cite{Sagonas2013_300Face} dataset contains 4437 images with 68 landmark annotations. AFLW \cite{AFLW2011} contains 24386 images with 21 landmark annotations, COFW \cite{Burgos2013COFW} contains 1852 images with 29 landmark annotations, and WFLW \cite{Wu2018WFLW} have 10000 images with 98 landmark annotations.

Several iris-landmark-detection methods \cite{choi2019accurate,lee2020deep,ablavatski2020real} uses cropped single eye regions from facial-landmark-detection results. Choi et al. \cite{choi2019accurate} proposed segmentation based eye center estimation. They cut out a rectangle region using the landmarks of the eye socket and eye corner from 68 points of landmarks \cite{kazemi2014one} before pupil segmentation. Ablabatski et al. \cite{ablavatski2020real} detects 5 points of iris landmarks from a $64 \times 64$ single eye region from facial-landmark detection \cite{bazarevsky2019blazeface} results.

Our assumption of NIR partial face image data is images under intense NIR illumination, such as CASIA-Iris-Distance \cite{casiairisdistance} and CASIA-Iris-M1 dataset \cite{CASIAIrisM1}. Since the modality of these images is different from the visible light data set, the pre-trained models of the above detection methods are insufficient to provide accuracy. In addition, there are currently hardly any public datasets with annotations for near-infrared light face images. Therefore, it is necessary to annotate facial bounding box and landmark annotations on existing datasets. However, these annotations are very costly for the task of eye detection.

\begin{figure*}[t]
\centering
\includegraphics[width=0.8\hsize]{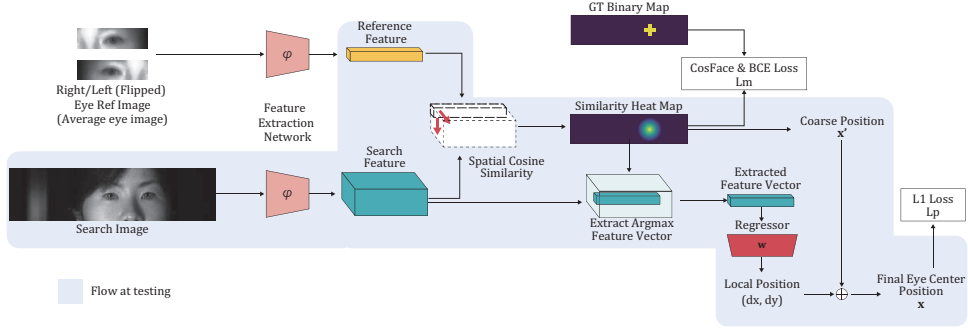}
\caption{
Framework of SiamEDP. Reference and search images are fed into same feature extractor. Heat map is generated by calculating spatial cosine similarity between extracted features. Networks are optimized by the binary cross entropy (BCE) loss with CosFace as shown in Eq. \ref{eq:loss}. Eye feature vector is extracted from similarity heat map, and regressor estimates fine position in eye feature. Blue region shows testing framework. Detector computes spatial cosine similarity between extracted search and pre-extracted reference features. Reference feature is saved as trained parameter.
}
\label{fig:flow}
\end{figure*}

\subsection{Direct Eye Detection Methods}
Methods have been proposed to detect eyes directly from input images using CNN-based object detection \cite{ren2015faster,redmon2018yolov3} without face and landmark detection. These methods, called generic object-detection methods \cite{liu2020deep}, detect and classify objects of different scales and classes.Faster R-CNN \cite{ren2015faster} is one such method. It generates a feature map from an input image by using convolutional layers and estimates regions of interest (ROIs) with high objectness using a region proposal network. Then, a fully connected layer outputs the object class and bounding box of the ROIs. Nasaif et al. \cite{nsaif2021frcnn} proposed FRCNN-GNB for eye detection. It uses Faster R-CNN \cite{ren2015faster} to detect the initial eye regions then applies Gabor filters and a naïve Bayes model for the final eye detection. Generic object-detection methods tend to have weak discriminability against similar classes such as right and left eye class, and tend to increase processing time due to the large size of the backbone to handle multiple classes and multiple scales.

We focused on object tracking methods for fast detection. Methods based on correlation filters or Siamese networks have been proposed and are often used in the Visual Object Tracking (VOT) challenge \cite{VOT_TPAMI}. Certain eye detectors using correlation filters have been proposed \cite{bolme2009average,araujo2014fast}. Araujo et al. \cite{araujo2014fast} proposed a correlation-filter-based method in the pixel domain. They use cosine similarity during training to avoid the values outside the interval $[0, 1]$. On the other hand, there is no eye detection method using Siamese network. The well-known object tracking method, SiamFC\cite{SiamFC} has achieved high performance and speed in the VOT challenge. It extracts features from input reference and search images by using common networks. A similarity score map is then generated using the correlation of the extracted reference and search features. SiamFC detects the position of a reference object in a search image by thresholding the similarity heat map.

\section{Proposed Method}
\subsection{Basic Idea}
We consider SiamFC \cite{SiamFC} as two-class classifier that determines whether a subregion of a search feature is a tracking target. We considered applying cosine-margin-based loss, which is recently used in metric learning. It is a method for showing higher classification performance by embedding features into a hypersphere and providing a margin for the same class on training. We apply CosFace \cite{wang2018cosface} to 2D convolution to improve classification performance. The position of eye in the heat map is rough because the resolution of heat map is reduced by the stride setting of feature extraction CNN architecture. Therefore, we considered a method to obtain a fine eye position using the coarsely obtained eye positions. We designed the spatial size of the search feature to be equal to the size of the heat map, and there is a feature vector corresponding to one pixel in the heat map. We assume a feature vector with the highest similarity includes the information of eye, and directly obtain the eye center coordinates by regressing on the feature vector. This enables high-speed and highly accurate eye detection.

\begin{table*}[t]
\centering
\small
\caption{Feature extraction network architecture}
\label{tb:network}
\begin{tabular}{lcccccc}
\hline 
&input&layer1&layer2&layer3&layer4&\\
\hline \\
convlayer&
$
\begin{bmatrix}
3 \times 3 ,32\\
\end{bmatrix}
$
&
$
\begin{bmatrix}
3 \times 3 ,32\\
3 \times 3 ,32\\
\end{bmatrix}
\times 2$
&
$
\begin{bmatrix}
3 \times 3 ,64\\
3 \times 3 ,64\\
\end{bmatrix}
\times 2$
&
$
\begin{bmatrix}
3 \times 3 ,128\\
3 \times 3 ,128\\
\end{bmatrix}
\times 2$
&
$
\begin{bmatrix}
3 \times 3 ,128\\
3 \times 3 ,128\\
\end{bmatrix}
\times 2$
\\
&stride 2&stride 2&stride 1&stride 2&stride 1\\

\hline
\end{tabular}
\end{table*}

\subsection{Framework}
We define reference image as $I_{ref}$ and search image as $I_{srch}$. The same CNN feature extractors $\phi$ calculate features from these two images. The features of $I_{ref}$ and $I_{srch}$ are defined as $f_{ref} := \phi(I_{ref}) \in \mathbb{R}^{m \times n \times c}$ and $F_{srch} := \phi(I_{srch}) \in \mathbb{R}^{M \times N \times c}$, respectively. The numbers of channels $c$ in $F_{srch}$ and $f_{ref}$ are equal, but each has a different spatial size. Let $F_{srch}[u] \in \mathbb{R}^{m \times n \times c}$ denote the partial region of size $m \times n \times c$ at spatial position $u \in U$ of feature $F_{srch}$. 

Figure \ref{fig:flow} shows the flow of SiamEDP. In training, $I_{ref}$ is fixed to a single average image of eyes, and the same image is flipped between the left and right eyes. We assume that $I_{srch}$ always includes both eyes and detection tasks for the left and right eyes are learned, respectively. In testing, the same $I_{ref}$ as in training is used to obtain both features $f_{refR}$ and $f_{refL}$ in advance. We define Q as the heat map generated by the cosine similarity between $F_{srch}$ and $f_{ref}$. We define $\mathbf{x'}$ as the spatial position where $argmax{Q}$, and extract the feature vector $f_{eye}$ such that $F_{srch}[\mathbf{x'}]\in \mathbb{R}^{1 \times 1 \times c}$. The local position $d\mathbf{x}$ of the eye center in $f_{eye}$ is obtained by regression on $f_{eye}$ and the final position is estimated using $\mathbf{x'}$ and $d\mathbf{x}$.

\subsection{Coarse to Fine Eye Center Estimation}
SiamEDP calculates the kernel-wise cosine similarity between $F_{srch}[u]$ and $f_{ref}$, and obtains $Q$. 
The edge of $F_{srch}$ is replicated before calculating the cosine similarity so that $Q$ has the same spatial size ($M, N$) as $F_{srch}$. The cosine similarity at spatial position $u$ on $Q$ can be calculated as 
\begin{equation}
\label{eq:map}
Q[u] = \cos{\theta_u}= \frac{f_{ref} \cdot F_{srch}[u]}{||f_{ref}||_2||F_{srch}[u]||_2} .
\end{equation}

The spatial position$\mathbf{x}'$ is highest in intensity in $Q$ as $\mathbf{x}' = [x',y']^T = argmax{Q}$. The feature $f_{eye}$ representing the eye is denoted as 
\begin{equation}
\label{eq:fea}
f_{eye} = F_{srch}[\mathbf{x}'] \in \mathbb{R}^{c}.
\end{equation}
A local eye center position $d\mathbf{x} = [dx, dy]$ in the feature block $f_{eye}$ is obtained by linear regression using weight parameters $\mathbf{w} \in \mathbb{R}^{2 \times c}$ as
\begin{equation}
\label{eq:reg}
d\mathbf{x} = 
\begin{bmatrix}
dx \\
dy
\end{bmatrix}
= \mathbf{w}f.
\end{equation}
Using $\mathbf{x}'$, $\mathbf{x}$ and the size ratio of the input image to the output similarity map $\alpha$, the final eye position coordinates $\mathbf{x}$ for the input image are obtained by $\mathbf{x} = \alpha(\mathbf{x}' + d\mathbf{x})$.

\subsection{Loss Function}
We define two types of loss functions. One is the loss $L_s$ for the similarity map, and the other is the loss $L_p$ for the eye center coordinates. We design a loss function on the basis of binary cross entropy (BCE) with CosFace \cite{wang2018cosface} to accurately classify one side of the eye (two classes) from others. CosFace has a margin parameter $m$ applied only to the positive label locations and a scale parameter $s$. These parameters enables cosine decision margin between classes. CosFace is expected to separate the feature distribution of the right-eye class from that of the left-eye class.
The loss function $L_s$ based on the BCE is given by
\begin{eqnarray}
\label{eq:loss}
L_s = - \frac{1}{|U|} \sum_{u \in U} \{y_u {\rm log}\frac{e^{s(Q_u - m)}}{ e^{s(Q_u - m)} + \Sigma_{t \neq u}{e^{s Q_t} }} \nonumber \\
+ (1-y_u) {\rm log}\frac{e^{s Q_u}}{ e^{s(Q_u - m)} + \Sigma_{t \neq u}{ e^{s Q_t} }} \}
.
\end{eqnarray}

The $L_p$ for fine eye center position is the L1 norm expressed as
\begin{equation}
\label{eq:lp}
L_p = \Sigma_{p \in {\mathbf x}}b\cdot|p-\hat{p}|\cdot \frac{1}{\alpha},
\end{equation}
where $\alpha$ denotes the scale ratio between the input image and similarity heat map. $b$ denotes a mask to avoid calculating regression losses for features without eye information due to incorrect heat-map predictions and defined as follows.
\begin{equation}
\label{eq:mask}
b = 
\left \{
\begin{aligned}
1, \ && if \ ||\mathbf{x}'-\mathbf{\hat{x}}'||_2<2
\\
0, \ && otherwize\ 
\end{aligned}
\right.
,
\end{equation}
The above $L_s$ and $L_p$ are combined into a loss function $L$ using the weights $\beta,\gamma$ as follows. We use a sum of loss calculated from the right and left eyes.
\begin{equation}
L = \beta L_s + \gamma L_p. 
\end{equation}

\begin{table}
\centering
\caption{Dataset}
\label{tb:Dataset}
\scalebox{0.7}[0.9]{
\begin{tabular}{lcl}
\hline 
Name& Number & Training\\
\hline
CASIA-Iris-Distance \cite{casiairisdistance} & 2567 & 60\% in \S\ref{sec:AS}, \S\ref{sec:LD}  \\
& & 50\% in \S\ref{sec:OD}  \\
CASIA-Iris-M1-S1 \cite{CASIAIrisM1} & 1400 & 60\% in \S\ref{sec:AS}, \S\ref{sec:LD} \\
& & 50\% in \S\ref{sec:OD}  \\
CASIA-Iris-M1-S2 \cite{CASIAIrisM1} & 6000 & 0\%\\
CASIA-Iris-M1-S3 \cite{CASIAIrisM1} & 3600 & 0\%\\
\hline
\end{tabular}
}
\end{table}

\begin{table*}
\centering
\caption{Ablation Study}
\label{tb:abStudy}
\begin{tabular}{llccccccc}
\hline
ref image&norm&params (s,m)&$e \leq 0.05$&$e \leq 0.1$&$e \leq 0.15$&$e \leq 0.2$&$e \leq 0.25$&$e \leq 0.5$\\
\hline
random&&&0.0001&0.0006&0.0037&0.0126&0.0287&0.2720\\
random&norm&(1,0)&0.4048&0.4098&0.4108&0.4122&0.4135&0.4147\\
random&norm&(30,0)&0.9850&0.9953&0.9964&0.9969&0.9972&0.9976\\
random&norm&(30,0.1)&0.9782&0.9947&0.9975&0.9983&0.9985&0.9990\\
random&norm&(30,0.2)&0.9839&0.9956&0.9981&{\bf 0.9991}&{\bf 0.9994}&{\bf 0.9995}\\
random&norm&(30,0.3)&0.9784&0.9936&0.9974&0.9988&0.9992&0.9994\\
fixed avg image&&&0.0000&0.0005&0.0023&0.0087&0.0272&0.1465\\
fixed avg image&norm&(1,0)&0.5171&0.6351&0.7062&0.7622&0.8181&0.9702\\
fixed avg image&norm&(30,0)&0.9866&0.9929&0.9940&0.9943&0.9947&0.9948\\
fixed avg image&norm&(30,0.1)&{\bf 0.9899}&{\bf 0.9975}&{\bf 0.9985}&0.9987&0.9988&0.9989\\
fixed avg image&norm&(30,0.2)&0.9776&0.9898&0.9917&0.9921&0.9924&0.9927\\
fixed avg image&norm&(30,0.3)&0.9798&0.9905&0.9918&0.9920&0.9922&0.9922\\
without ref image&norm&(30,0.1)&0.7509&0.7831&0.7876&0.7899&0.7919&0.8018\\
\hline
\end{tabular}
\end{table*}

\begin{table*}
\centering
\caption{Combinations of comparison methods. We evaluated 9 combinations from 4 face SDKs. Some facial-landmark-detection methods do not detect eye centers; therefore, we used mean of points around eye.}
\label{tb:SDK}
\scalebox{0.6}[0.7]{
\begin{tabular}{lllll}
\hline 
Name & ToolBox & Face Detection&Landmark& Eye-center Evaluation\\
\hline 
Dlib-5points & Dlib \cite{king2009dlib} & \cite{dalal2005histograms} & 5 points, \cite{kazemi2014one} & 
mean of 2 endpoints of eye\\
Dlib-68points & Dlib \cite{king2009dlib} & \cite{dalal2005histograms} & 68 points, \cite{kazemi2014one} & 
mean of 6 points around eye\\
Dlib-cnn-5points & Dlib \cite{king2009dlib}& \cite{king2015max} & 5 points, \cite{kazemi2014one} &
mean of 2 endpoints of eye\\
Dlib-cnn-68points & Dlib \cite{king2009dlib}& \cite{king2015max} & 68 points, \cite{kazemi2014one} & 
mean of 6 points around eye\\
FaceX-Zoo-mask & FaceX-Zoo \cite{wang2021facex} & RetinaFace \cite{deng2020retinaface},  & 106 points, PFLD \cite{guo2019pfld} & point of eye center
\\
&&(trained mask data)
\\
FaceX-Zoo-non-mask & FaceX-Zoo \cite{wang2021facex}& RetinaFace \cite{deng2020retinaface} & 106 points, PFLD \cite{guo2019pfld}&
point of eye center
\\
Mediapipe-iris & Mediapipe \cite{lugaresi2019mediapipe} & BlazeFace \cite{bazarevsky2019blazeface} & 468 points, FaceMesh \cite{kartynnik2019real} & point of iris center
\\
&&& + iris 5 points, \cite{ablavatski2020real} & \\
Mediapipe& Mediapipe \cite{lugaresi2019mediapipe} & BlazeFace \cite{bazarevsky2019blazeface} & 468 points, FaceMesh \cite{kartynnik2019real} & 
mean of 100 points around eye
\\
OpenFace2.0& OpenFace2.0 \cite{baltrusaitis2018openface} & \cite{xiang2017joint} & 68 face points, CE-CLM \cite{zadeh2017convolutional}  & 
mean of 8 points around pupil\\
&&&+ 56 eye points, \cite{wood2015rendering} &
\\
\hline
\end{tabular}
}
\end{table*}

\section{Experiments and Results}
We present three experiments we conducted evaluate the performance of SiamEDP. The first was an ablation study to confirm the contributions of SiamEDP (SiamFC, cosine similarity and CosFace). The second was a comparison between SiamEDP and current facial-landmark-detection methods. The third was a comparison between SiamEDP and generic object-detection methods. We considered SiamEDP is preprocessing for iris and pupil segmentations, so we did not compare segmentation methods.

We used a single network architecture for all three experiments. The base network for SiamFC was modified from ResNet \cite{he2016deep} as shown in Table \ref{tb:network}. The differences from the original ResNet are single channel input and the number of layers. Each layer contained convolution, batch normalization and rectified-linear-unit (ReLU) activation. For model training, we used stochastic gradient descent (SGD) as the optimizer. The learning rate was 0.1 on the first epoch and switched to 0.01 from the second epoch. The weight decay was 0.0001. The batch size was 16 for each iteration, and the total number of iterations was 60,000. 

We used four iris datasets, CASIA-Iris-Distance \cite{casiairisdistance}, CASIA-Iris-M1-S1\cite{CASIAIrisM1}, CASIA-Iris-M1-S2\cite{CASIAIrisM1}, and CASIA-Iris-M1-S3 \cite{CASIAIrisM1} described in Table. \ref{tb:Dataset}. We manually annotated eye center points on all images. The input image was scaled down from the original size, with the resolution of the iris diameter at about 10 pixels. Therefore, the images of CASIA-Iris-M1-S1 were rescaled to 1/10, the others were rescaled to 1/16. Parameter $\alpha$ is the scale ratio between the resized input image and similarity heat map and is $8$ because of the network stride in Table \ref{tb:network}. The ground truth heat map is a binary map labelled on the eye-center pixel and its four neighboring pixels.

The evaluation metric was the root-mean-square error (RMSE) of the eye-center position or the relative error considering both eyes and expressed as 
\begin{equation}
\label{eq:eyeError}
E = \frac{max{(d_l,d_r)}}{d},
\end{equation}
where $d$ is the Euclidean distance between the left- and right-eye centers, where $d_l$ and $d_r$ are the RMSEs of the right- and left-eye centers, respectively.

\begin{table*}
\centering
\caption{Comparison of SiamEDP and facial-landmark-detection methods on CASIA-Iris-M1-S2. The face detection column is the success rate of the face detection API returned bounding box.}
\label{tb:landmarkss2}
\scalebox{0.9}[0.9]{
\begin{tabular}{l|c|c}
\hline 

& Eye error & RMSE \\
\hline 
\begin{tabular}{l}
\\
Dlib-5points\\
Dlib-68points\\
Dlib-cnn-5points\\
Dlib-cnn-68points\\
FaceX-Zoo-mask\\
FaceX-Zoo-non-mask\\
Mediapipe-iris\\
Mediapipe\\
OpenFace2.0\\
SiamEDP (proposed)\\
\end{tabular}
&
\begin{tabular}{cccc}
Face detection & $\leq 0.05$ &$\leq 0.1$&$\leq 0.25$\\
0.0097&0.0307&0.3655&0.7467\\
0.0097&0.1193&0.5293&0.7240\\
0.3962&0.0940&0.5337&0.8470\\
0.3962&0.2323&0.6662&0.8290\\
0.8212&0.0467&0.3068&0.6348\\
0.6043&0.0463&0.3002&0.5603\\
0.5688&0.1427&0.4523&0.5520\\
0.5688&0.0855&0.4805&0.5667\\
0.0202&0.0137&0.0192&0.0198\\
&{\bf 0.9883}&{\bf 0.9970}&{\bf 0.9980}\\
\end{tabular}

&
\begin{tabular}{cccc}
$\leq 5$&$\leq 10$&$\leq 15$&$\leq 20$\\
0.5505&0.7686&0.9018&0.9664\\
0.6082&0.7365&0.8644&0.9552\\
0.6786&0.8514&0.9287&0.9768\\
0.7377&0.8398&0.9067&0.9706\\
0.5248&0.7034&0.7638&0.7902\\
0.4662&0.5778&0.5979&0.6017\\
0.4958&0.5494&0.5642&0.5674\\
0.5332&0.5667&0.5680&0.5682\\
0.0200&0.0201&0.0202&0.0202\\
{\bf 0.9983}&{\bf 0.9988}&{\bf 0.9991}&{\bf 0.9991}\\
\end{tabular}
\\
\hline
\end{tabular}
}
\end{table*}

\begin{table*}
\centering
\caption{Comparison of SiamEDP and facial-landmark-detection methods on CASIA-Iris-M1-S3. The face detection column is the success rate of the face detection API returned bounding box.}
\label{tb:landmarkss3}
\scalebox{0.9}[0.9]{
\begin{tabular}{l|c|c}
\hline 

& Eye error & RMSE \\
\hline 
\begin{tabular}{l}
\\
Dlib-5points\\
Dlib-68points\\
Dlib-cnn-5points\\
Dlib-cnn-68points\\
FaceX-Zoo-mask\\
FaceX-Zoo-non-mask\\
Mediapipe-iris\\
Mediapipe\\
OpenFace2.0\\
SiamEDP (proposed)\\
\end{tabular}
&
\begin{tabular}{cccc}
Face detection & $\leq 0.05$ &$\leq 0.1$&$\leq 0.25$\\
0.3008&0.0464&0.2994&0.6569\\
0.3008&0.1869&0.3697&0.6003\\
0.7614&0.0547&0.3742&0.7828\\
0.7614&0.3214&0.6114&0.7914\\
0.9903&0.1917&0.8139&0.9819\\
0.9833&0.1906&0.8108&0.9758\\
0.9728&0.6142&0.9375&0.9714\\
0.9728&0.2000&0.8708&0.9714\\
0.8572&0.6489&0.8142&0.8428\\
&{\bf 0.9894}&{\bf 0.9981}&{\bf 0.9983}\\
\end{tabular}

&
\begin{tabular}{cccc}
$\leq 5$&$\leq 10$&$\leq 15$&$\leq 20$\\
0.5400&0.7776&0.8810&0.9500\\
0.4931&0.6440&0.7928&0.9175\\
0.6356&0.8636&0.9267&0.9676\\
0.7163&0.8238&0.8967&0.9489\\
0.8983&0.9812&0.9878&0.9892\\
0.8939&0.9751&0.9811&0.9824\\
0.9560&0.9717&0.9722&0.9722\\
0.9242&0.9714&0.9722&0.9722\\
0.8282&0.8438&0.8499&0.8549\\
{\bf 0.9988}&{\bf 0.9992}&{\bf 0.9993}&{\bf 0.9994}\\
\end{tabular}
\\
\hline
\end{tabular}
}
\end{table*}

\subsection{Ablation Study}
\label{sec:AS}
In this experiment, we evaluated the performance of selecting reference images, effectiveness of CosFace, and that of the Siamese network. We selected CASIA-Iris-Distance and CASIA-M1-S1 as training datasets and used 60\% of each training data set for training. 

We first evaluated the effectiveness of the CosFace parameters $(s,m)$ and feature normalization and estimated suitable parameters. For training without reference images, we set the best parameters in the evaluation of CosFace. We trained all models three times, each with the same parameters and same training data, then averaged the results. The evaluation was done using CASIA-M1-S2 and CASIA-M1-S3 datasets, which were not used for training.

We then evaluated the following three methods for selecting reference images: 
\begin{enumerate}
   \item {\bf random}. 10\% of the images in the training dataset is selected as reference image data, and eyes were randomly selected from them and used as reference images during training. In the evaluation, we used the average of the reference features in a batch in the end of training as a reference feature.
   \item {\bf fixed avg image}. The histogram-stretched image averaged over 128 reference images is used as the reference image. The reference average image is fixed during the training and testing. 
   \item {\bf without ref image}. A heat map is learned directly from a search feature without using the reference image. Channel 2 and 1×1 convolution is executed on the search features to directly output the right- and left-eye heat maps. We applied CosFace to this method and selected parameters from the other methods.
\end{enumerate}

The results are listed in Table \ref{tb:abStudy}. There was no significant difference between random reference and avg reference, and both methods significantly decreased in accuracy when the features were not normalized. We also found significant performance differences depending on the presence or absence of $s$ (same as NormFace \cite{wang2017normface}). Without reference images, the learning was not stable, resulting in lower accuracy. 

\begin{figure*}[t]
\centering
\includegraphics[width=0.8\hsize]{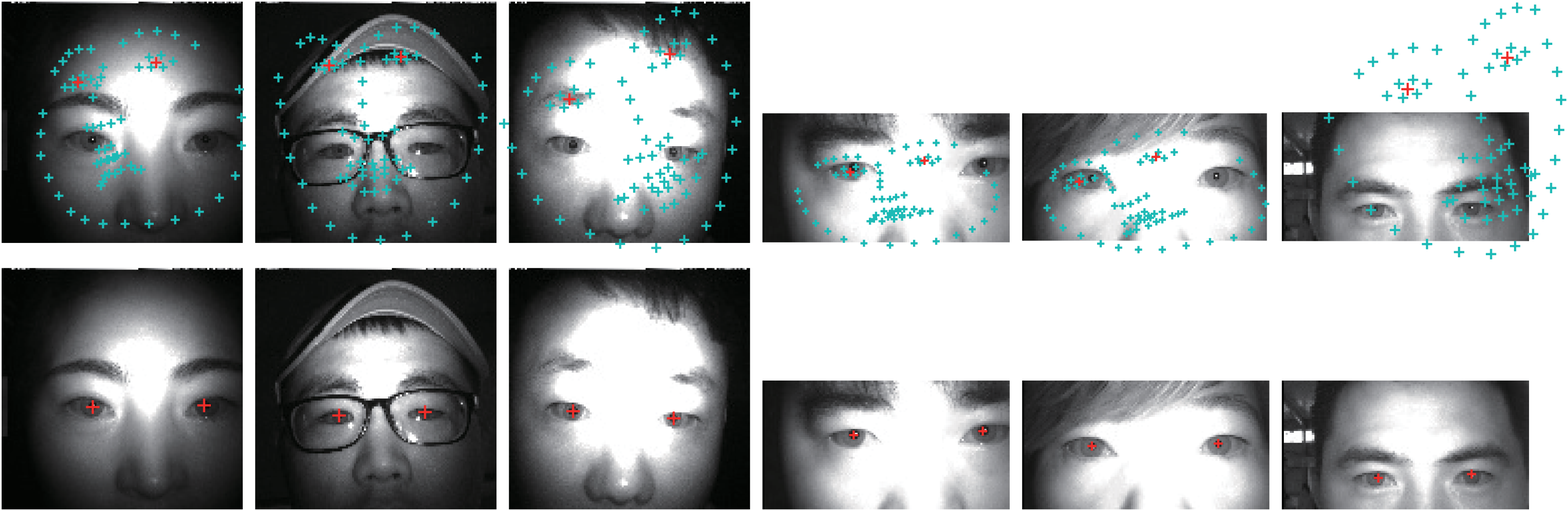}
\caption{Results of Dlib (use CNN based detection and 68 points detection) and SiamEDP. The left three columns are the results using the images from CASIA-Iris-M1-S3 \cite{CASIAIrisM1}. Right three columns are results using the images from CASIA-Iris-M1-S2 \cite{CASIAIrisM1}.}
\label{fig:DLibs}
\end{figure*}

\begin{figure}[h]
\centering
\includegraphics[width=0.85\hsize]{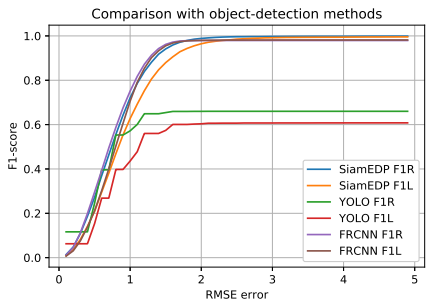}
\caption{Comparisons with generic object detection methods. We evaluate F1 scores of each eyes per RMSE.}
\label{fig:objF1Scores}
\end{figure}

\subsection{Landmark Detection}
\label{sec:LD}
In the next experiment, we evaluated pre-trained face-detection and facial-landmark-detection models, which are publicly available as software development kits (SDKs) for face detection and recognition, and compared them using relative eye error and RMSE using eye-center points or the average of landmarks around the eye. We evaluated possible combinations of the methods in each of the four SDKs: Dlib \cite{king2009dlib}, FaceXZoo \cite{wang2021facex}, Mediapipe \cite{lugaresi2019mediapipe}, and OpenFace \cite{baltrusaitis2018openface}. The specifications of most SDKs first requires calculating the bounding box by face detection, then inputing the image and bounding box region to landmark detection. Since many SDKs did not detect faces in the CASIA datasets and did not output bounding boxes from detection modules, we recorded the success rate of detecting at least one bounding box and carried out landmark detection when a bounding box was detected. As an exception, when the evaluation using Dlib failed to detect faces, the bounding box was set as the entire image area and input to landmark detection because we assumed could be optimized from the initial points on the partial face image. Several facial-landmark-detection methods output only landmarks around the eyes, so the average value of the points around the eyes is output as the eye center. The combinations of each method and calculate eye-center positions are listed in Table \ref{tb:SDK}.

The results of CASIA-Iris-M1-S2 are listed in Table \ref{tb:landmarkss2} and those of CASIA-Iris-M1-S3 are listed in Table \ref{tb:landmarkss3}. The accuracy of face detection for CASIA-Iris-M1-S2 is decreased because most of the images are partial images of the face (from the nose up). Since landmark detection with Dlib involves optimization by placing initial points, it can achieve 90\% of images with RMSE less than 15 to some extent even if applied directly to the image but requires a large margin when cropped to a single eye. CASIA-Iris-M1-S3 includes the entire face, which has shown higher performance on SDKs than CASIA-Iris-M1-S2, but it is less accurate than SiamEDP in Eye relative Error and RMSE. Fig. \ref{fig:DLibs} shows detection examples with Dlib.

We evaluated processing times averaged 10 times in a CPU environment on CASIA-Iris-M1-S3 scaled down to 120 $\times$ 120 pixels. SiamEDP was 14 msec, FaceXZoo was 32.5 msec/33 msec depending if detection engine for mask was used, mediapipe facial-landmark detection was 14 msec and 35 msec with iris-landmark detection, Dlib with CNN face detection is 219 msec with 68-point detection, 218 msec with 5-point detection, Dlib with not-CNN-based detection is 10 msec with 68-point detection and 9 msec with 10 msec 5-point detection. We were unable to measure exact execution times when using OpenFace because we evaluated using a built executable file. Therefore, SiamEDP fast and the most accurate.

\subsection{Object Detection}
\label{sec:OD}
We compared SiamEDP with the major generic object detection methods FRCNN \cite{ren2015faster} and YOLOv5 (a PyTorch implementation of YOLOv4 \cite{yolov4}). We re-trained the YOLO and FRCNN models to detect both eyes from partial face images. The YOLO models were trained in 70 epochs, while FRCNN was trained in 5 epochs because the training time with FRCNN is much longer than the other methods. 

The training datasets included CASIA-Iris-Distance and CASIA-Iris-M1-S1. We trained 50\% of the images from the training domain datasets. The other 50\% of the training domain datasets was used in the evaluation (trained domain). The un-training domain datasets included CASIA-Iris-M1-S2 and CASIA-Iris-M1-S3. All un-training domain datasets (untrained domain) were used for evaluation. We labeled a 16 $\times$ 32 pixels bounding box centered on the eye center with two classes, i.e., right eye and left eye, for YOLO and FRCNN.

We compared the three methods using the F1 score per RMSE to evaluate the discriminative performance between right- and left-eye classes. Since it assumed with SiamEDP that both eyes are always included in an input image, a false positive (FP) is always equal to a false negative (FN). This assumption makes precision equal to the recall for SiamEDP. In generic object detection, the number of detection targets is unlimited, so YOLO and FRCNN may detect more targets than expected (the FN may differ from the FP). Thus, we needed to evaluate them on the basis of the F1 score instead of accuracy. 

The results are shown in Figure \ref{fig:objF1Scores}. 
The results of YOLO indicate that the F1 scores converged around 0.7, indicating low discriminability between the right and left eyes. FRCNN had the highest accuracy in regions where RMSE was small. The results of SiamEDP are converged to the highest F1 score. The execution speed of SiamEDP was 11m sec, that of FRCNN was 1970 msec, and that of YOLO was 87 msec on average of 10 CPU runs for a $123 \times 96$ pixels of image, with SiamEDP having the best results.

\section{Conclusion}
We proposed a fast eye-detection method that is based on a Siamese network for NIR partial face images. By using the Siamese network and cosine-margin-based loss function, a shallow network was able to detect the left- and right-eye centers with high accuracy. Experimental results indicate that the Siamese network and CosFace is effective in achieving high-speed and high-accuracy detecion in a CPU environment compared with current facial-landmark-detection methods and generic object-detection methods.

\bibliographystyle{apalike}
{\small
\bibliography{refs}}

\end{document}